\documentclass[runningheads]{llncs}
\usepackage{template}
\usepackage{abbrv}

\usepackage{graphicx}
\usepackage{booktabs}
\usepackage{multirow}
\usepackage{subcaption}
\usepackage{algorithm2e}
\usepackage[accsupp]{axessibility} 
\usepackage{hyperref}
\usepackage{orcidlink}
\usepackage{arydshln}   
\usepackage{adjustbox}
\usepackage[accsupp]{axessibility}  

\usepackage[table]{xcolor}
\usepackage{arydshln} 

\definecolor{bestbg}{RGB}{255, 230, 153}   
\definecolor{secondbg}{RGB}{255, 242, 204} 
\usepackage{hyperref}
\usepackage{orcidlink}
\usepackage{enumitem}

\begin{document}
\title{
\texorpdfstring{Poppy: \underline{Po}larization-based \underline{P}lug-and-\underline{P}la\underline{\smash{y}} Guidance for Enhancing \\ Monocular Normal Estimation}%
{Poppy: Polarization-based Plug-and-Play Guidance for Enhancing Monocular Normal Estimation}%
}

\titlerunning{Poppy}

\author{Irene Kim\inst{1}\orcidlink{0009-0001-6917-6894} \and
Sai Tanmay Reddy Chakkera\inst{1}\orcidlink{0009-0002-9695-8784} \and
Alexandros Graikos\inst{1}\orcidlink{0009-0004-5170-1074} \and
Dimitris Samaras \inst{1}\orcidlink{0000-0002-1373-0294} \and
Akshat Dave\inst{1}\orcidlink{0000-0003-0560-632X}}

\authorrunning{I. Kim et al.}
\institute{Stony Brook University
\email{\{irnkim,schakkera,agraikos,samaras,dave\}@cs.stonybrook.edu}}

\maketitle

\begin{figure}[ht]
  \centering
  \includegraphics[width=\textwidth]{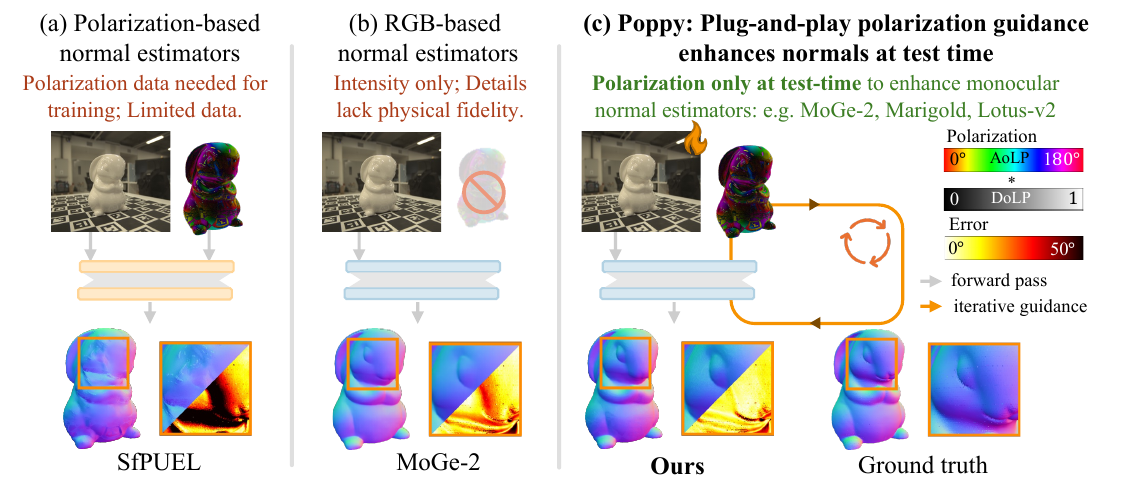}
  \caption{
    \textbf{Test-time polarization guidance to enhance normal estimation.}
    (a) Polarization-based feed-forward models have limited generalizability due to the scarcity of polarization--normal training pairs.
    (b) RGB-only monocular normal estimators produce oversmoothed or hallucinated details on challenging surfaces -- normals of the textureless bunny object appear flatter than ground truth.
    (c) Poppy introduces polarization guidance into pretrained RGB-only models at test time -- improving normal accuracy without retraining.
  }
  \label{fig:teaser}
\end{figure}
\begin{abstract}
  Monocular surface normal estimators trained on large-scale RGB-normal data often perform poorly in the edge cases of reflective, textureless, and dark surfaces. Polarization encodes surface orientation independently of texture and albedo, offering a physics-based complement for these cases. Existing polarization methods, however, require multi-view capture or specialized training data, limiting generalization. We introduce Poppy, a training-free framework that refines normals from any frozen RGB backbone using single-shot polarization measurements at test time. Keeping backbone weights frozen, Poppy optimizes per-pixel offsets to the input RGB and output normal along with a learned reflectance decomposition. A differentiable rendering layer converts the refined normals into polarization predictions and penalizes mismatches with the observed signal. Across seven benchmarks and three backbone architectures (diffusion, flow, and feed-forward), Poppy reduces mean angular error by 23--26\% on synthetic data and 6--16\% on real data. These results show that guiding learned RGB-based normal estimators with polarization cues at test time refines normals on challenging surfaces without retraining.
\end{abstract}

\section{Introduction}
\label{sec:intro}
\noindent Surface normal estimation is a core problem in computer vision, with applications in robotics, augmented reality, scene understanding, and 3D reconstruction. Monocular normal estimation aims to recover per-pixel 3D surface orientation from a single RGB image -- an inherently ill-posed task because many surface normals can produce the same 2D appearance. Modern learning-based normal estimators reduce this ambiguity by learning geometric priors from large-scale paired RGB and normal datasets \cite{dsine,stablenormal,marigold,moge,He2025Lotus2AG}.

However, state-of-the-art monocular estimators fail on three common surface types where RGB cues alone are unreliable: 1) highly reflective surfaces, where view-dependent highlights are misinterpreted as geometry; 2) textureless regions, which lack spatial variation and produce oversmoothed normals; and 3) dark objects, where low signal-to-noise ratio degrades predictions. RGB-normal datasets often under-represent these surfaces, compounding these errors (Fig.~\ref{fig:teaser}(b)). Thus, in this paper, in addition to RGB, we capture the polarization of light,  only at test time, to refine normal estimation for these challenging scenarios. 

Polarization provides a physically grounded complement to RGB.
When unpolarized light reflects from a surface, it becomes partially polarized depending on surface orientation and material -- making polarization informative for the failure cases above.
Classical shape-from-polarization (SfP), however, is ill-posed: the azimuthal flip ($\pi$) ambiguity and the diffuse--specular ($\pi/2$) ambiguity together yield four candidate normals from a single measurement.
Prior classical methods resolve these ambiguities by acquiring additional measurements -- varying the illumination~\cite{polarpphoto,ATKINSON2017158,shadingpolar,ding2021polarimetric}, capturing multiple views~\cite{pandora,neisf,nersp,pisr}, or capturing additional modalities~\cite{polar3d,refr} -- increasing the capture burden beyond a single snapshot.

Learning-based SfP methods~\cite{sfpuel,dsfp,sfpwild} aim to retain single-snapshot capture by leveraging data-driven priors learned from curated polarization--normal training pairs.
These approaches, however, require paired polarimetric training data that is expensive to collect and narrow in scene diversity, which limits generalization to new materials and environments (Fig.~\ref{fig:teaser}a).
Neither classical nor learned SfP methods exploit the strong geometric priors already embedded in modern RGB normal estimators.

To bridge this gap, we introduce Poppy, a training-free framework that guides any differentiable RGB normal estimator with physics-based polarization constraints at test time (Fig.~\ref{fig:teaser}c). The weights of the normal estimator backbone are kept frozen. Poppy instead introduces learnable parameters before and after the network so that the predicted normals align with the observed polarization signal.
Poppy is plug-and-play: it applies to diffusion-based \cite{marigold}, flow-based \cite{He2025Lotus2AG}, and feed-forward \cite{moge} backbones without retraining or architectural changes. The pretrained backbone initializes normals using global scene structure from RGB as orientation constraints, resolving the azimuthal ambiguity; polarization guidance then corrects normal estimation errors that RGB cues alone miss.

Concretely, Poppy adds a learnable per-pixel image offset to the network input and a learnable per-pixel normal offset to the network output.
The image offset acts as a global perturbation: input changes propagate through the backbone, steering the predicted geometry at a scene-wide scale.
The normal offset refines high-frequency detail that the backbone misses.
A differentiable polarization rendering layer converts the refined normals and a learned specular radiance into Stokes vectors, a representation of the polarization state, resolving the remaining diffuse--specular ambiguity.
The loss between the predicted and measured Stokes vectors is minimized by optimizing the image offset, normal offset, and specular radiance maps.

We evaluate Poppy extensively across seven benchmarks -- SfPUEL~\cite{sfpuel} NeRSP~\cite{nersp}, NeISF~\cite{neisf}, and DeepPol~\cite{deeppol} (synthetic); SfPUEL, NeRSP, and PISR~\cite{pisr} (real) -- covering most publicly available polarization--normal datasets. Poppy enables consistent improvements in these benchmarks across three backbone architectures: Marigold, Lotus-v2, and MoGe-2 -- with 23--26\% reduction in mean angular error (MAE) for synthetic data and 6--16\% reduction on real data.  
Poppy demonstrates improved normal quality on challenging materials, including reflective, textureless, and low albedo surfaces. The refined normals also improve downstream 3D mesh quality when used to aid an existing multi-view reconstruction method. The learned per-pixel specular radiance enables decomposition of diffuse and specular Stokes.

Our contributions are:
\begin{enumerate}
  \item A training-free guidance framework that refines normals from RGB-based monocular estimators using polarization measurements at test time.
  \item A plug-and-play guidance mechanism with learnable per-pixel image and normal offsets for global-then-local refinement of any differentiable backbone, together with a learned per-pixel specular radiance for handling diffuse--specular ambiguities.
  \item Consistent improvements across three backbone architectures on seven benchmarks, demonstrating gains on reflective, textureless, and low-SNR surfaces that are typically challenging for normal estimation.
\end{enumerate}

\section{Related work}
\paragraph{Monocular surface normal estimation} methods predict per-pixel surface orientation from a single RGB image.
Early discriminative approaches~\cite{eigen2015predictingdepthsurfacenormals,9184024,do2022surfacenormalestimationtilted} train convolutional networks on large-scale supervision, with Omnidata~\cite{eftekhar2021omnidatascalablepipelinemaking} showing the benefit of multi-task pretraining.
More recent discriminative methods incorporate stronger priors: DSINE~\cite{dsine} encodes per-pixel camera geometry, and MoGe~\cite{moge} regresses affine-invariant point maps via a transformer backbone.
Diffusion-based methods~\cite{marigold,Garcia2024FineTuningID,Fu2024GeoWizardUT,Long2023Wonder3DSI} repurpose pretrained generative models for geometry estimation, with Marigold~\cite{marigold}, Lotus~\cite{lotus}, StableNormal~\cite{stablenormal}, and GenPercept~\cite{genpercept} among the most prominent.
Despite strong average-case performance, all of these methods rely solely on RGB appearance cues and degrade on reflective, textureless, and low-albedo surfaces where photometric shading signals are ambiguous.
Our work targets precisely these failure cases by introducing polarization-based physical constraints at test time.

\paragraph{Shape from Polarization} methods recover surface normals from reflected light using diffuse and height-from-polarization models~\cite{diffusepol,8456615,tozza2017lineardifferentialconstraintsphotopolarimetric}, specular and refractive cues~\cite{refr,polar3d,1238455,Morel2005PolarizationIA}, or joint shading--polarization formulations~\cite{Ichikawa_2021_CVPR,ATKINSON2017158,polarpphoto,shadingpolar}.
However, the polarization-to-normal mapping is inherently ambiguous due to $\pi$ and $\pi/2$ symmetries in the Fresnel equations.
Multi-view methods~\cite{atkinson2007twoviews,Cui2017PolarimetricMS,Yang_2018_CVPR,nersp,neisf,pisr} resolve this geometrically, while discriminative methods such as DeepSfP~\cite{dsfp} and SfPUEL~\cite{sfpuel} learn to predict normals directly from polarization images.
All of these approaches either require controlled capture, curated polarization training data, or multi-image acquisition.
Most closely related to our work, PPFT~\cite{ikemura2024robust} adapts a pretrained \emph{depth} model to polarization inputs via LoRA fine-tuning---requiring polarization training data and modifying model weights; in contrast, Poppy targets \emph{normals} and operates entirely at test time without retraining.

\paragraph{Test-time guidance} steers pretrained models at inference time rather than 
training specialized models on new data.
For diffusion models, guidance methods first imposed linear constraints by 
back-propagating a task-specific likelihood during the reverse diffusion, effectively 
steering generation towards measurement-consistent solutions \cite{chung2023dps}. More 
recent approaches have generalized this idea to arbitrary non-linear constraints 
\cite{yu2023freedom}, and faster, backpropagation-free sampling under constraints 
\cite{graikos2024fast}.
Guidance has been extensively used in depth-completion methods~\cite{viola2024marigolddc,
hyoseok2025zero,jeong2025test,talegaonkar2025repurposingmarigoldzeroshotmetric}, where 
sparse depth measurements or physics-based cues are used to guide pretrained monocular 
depth estimators at test time, as well as in medical image segmentation~\cite{chen2023vptta}, 
where image-specific visual prompts are optimized at inference to bridge cross-domain 
distribution shifts without modifying model weights.
Among the depth methods, Marigold-DC~\cite{viola2024marigolddc} is closest to our 
approach: it guides a diffusion-based depth model with sparse depth observations, whereas 
Poppy guides a normal estimation model with dense polarization measurements that encode 
shape through the Fresnel equations.

\section{Background}
\label{sec:polarization}
\subsection{Representing polarization}
\noindent Polarization characterizes the oscillation of light waves.
While natural illumination is often unpolarized, reflection from surfaces induces partial polarization that depends on surface orientation and material properties.
This makes polarization a physically grounded cue for geometry estimation.

\paragraph{Stokes vector.}
The polarization state of a light ray can be represented by the Stokes vector, $\mathbf{S} = [S_0, S_1, S_2, S_3]^\top$~\cite{collett2005field}.
$S_0$ represents total intensity.
$S_1$ measures the difference between horizontally and vertically polarized components.
$S_2$ measures the difference between components polarized at $45^\circ$ and $135^\circ$.
$S_3$ represents circular polarization.

A linear polarization camera captures  ($I$) through polarizers at multiple angles $(0^\circ, 45^\circ, 90^\circ, 135^\circ)$.
The linear Stokes parameters are computed as
\begin{equation}
  S_0 = I_{0} + I_{90}, \quad
  S_1 = I_{0} - I_{90}, \quad
  S_2 = I_{45} - I_{135}.
  \label{eq:stokes_linear}
\end{equation}
In most passive imaging scenarios, circular polarization is negligible compared to linear polarization.
We therefore restrict our formulation to linear polarization and use the reduced representation, 
$\mathbf{S} = [S_0, S_1, S_2]^\top.$

\paragraph{Degree and Angle of Polarization.}
The Degree of Linear Polarization (DoLP), denoted $\rho$, measures the fraction of light that is linearly polarized -- ranging from 0, for unpolarized light, to 1, for fully polarized light. DoLP can be obtained from the Stokes vector as 
$\rho = \frac{\sqrt{S_1^2 + S_2^2}}{S_0}.$
The Angle of Linear Polarization (AoLP), denoted $\phi$, describes the dominant polarization orientation. AoLP depends on the Stokes vector as 
$\phi = \frac{1}{2}\operatorname{arctan}(S_2, S_1)$.
Together, $(\rho, \phi)$ provide a compact representation of the linear polarization state.

\begin{figure}[ht]
  \centering
  \includegraphics[width=1\linewidth]{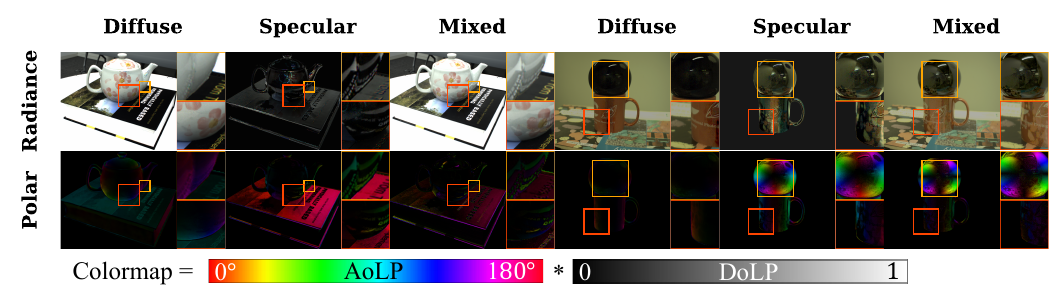}
  \caption{\textbf{Radiance decomposition.}
  The mixed radiance $S_0$ is decomposed into diffuse radiance $L_d$ and specular radiance $L_s$ from our method. The learned $L_s$ captures specular highlights and environment-dependent reflections (scaled 2$\times$ for clarity), while $L_d$ retains the object's intrinsic diffuse shading and texture. From these radiance components and the predicted normals, the polarization maps (AoLP$\times$DoLP) of the diffuse, specular, and combined components can be obtained.
  }
  \label{fig:radiance_decomposition}
\end{figure}

\paragraph{Diffuse and specular reflectance.}
When light reflects from a surface, it can be decomposed into diffuse and specular components. We provide an example decomposition learned from our method for visualization in Fig.~\ref{fig:radiance_decomposition}.
Diffuse reflection arises from subsurface scattering and typically produces weaker polarization, whereas specular reflection arises from surface reflection and follows Fresnel reflection laws~\cite{baek}.
Let $L_d$ and $L_s$ denote the diffuse and specular radiance components.
The observed Stokes vector is modeled as the sum of diffuse and specular contributions: 
$\mathbf{S} = S_{d} + S_{s}.$
Using the DoLP and AoLP of each component, the combined Stokes parameters can be written as~\cite{pandora,Li2023FoolingPV}
\begin{subequations}\label{eq:final_stokes}
  \begin{align}
    S_0 &= L_d + L_s \label{eq:final_stokes_s0}\\
    S_1 &= L_d \rho_d \cos(2\phi_d) + L_s \rho_s \cos(2\phi_s) \label{eq:final_stokes_s1}\\
    S_2 &= L_d \rho_d \sin(2\phi_d) + L_s \rho_s \sin(2\phi_s) \label{eq:final_stokes_s2} \; ,
  \end{align}
\end{subequations}
where $(\rho_{d/s},\phi_{d/s})$ represent diffuse/specular DoLP, AoLP respectively.

\subsection{Surface normals to Stokes}\label{sec:n2s}
\noindent Polarization provides a cue for surface orientation because both DoLP and AoLP depend on the surface normal relative to the viewing direction.

\paragraph{Spherical coordinate conversion.}
Denoting the surface normal as $n = [n_x, n_y, n_z]^\top$, we convert it into spherical coordinates $(\theta, \psi)$ relative to the viewing direction $v$, where the elevation angle $\theta = \arccos(n \cdot v)$ and azimuth angle $\psi = \operatorname{arctan}(n_y, n_x)$.

\paragraph{DoLP from surface normals.}
Under Fresnel reflection theory, the degree of polarization depends on the elevation angle $\theta$ and the refractive index $\eta$ (set to 1.5 per standard assumptions).

For diffuse reflection,
$
  \rho_d(\theta) =
  \frac{(\eta - 1/\eta)^2 \sin^2 \theta}
  {2 + 2\eta^2 - (\eta + 1/\eta)^2 \sin^2 \theta + 4 \cos \theta \sqrt{\eta^2 - \sin^2 \theta}},
$
and for specular reflection,
$
  \rho_s(\theta) =
  \frac{2\sin^2\theta \cos\theta \sqrt{\eta^2 - \sin^2\theta}}
  {\eta^2 - \sin^2\theta - \eta^2\sin^2\theta + 2\sin^4\theta}.
$

\paragraph{AoLP from surface normals.}
\label{sec:aolp_compents}
The angle of polarization is determined by the azimuth angle $\psi$.
For diffuse reflections,  $\phi_d = \psi$, and for specular reflections, $\phi_s = \psi + \pi/2$.
The orthogonality between these components reflects the phase difference induced by Fresnel reflection.

\paragraph{Stokes from surface normals.}
Normal-to-Stokes conversion is an inverse problem of SfP. Given a surface normal $n$ and the specular radiance $L_s$, we compute the diffuse and specular DoLP/AoLP $(\rho_{d/s}, \phi_{d/s})$ and diffuse radiance $L_d = S_0 - L_s$. Therefore, we can rewrite Eq.~\ref{eq:final_stokes} in terms of $n$ and $L_s$ and compute Stokes vector as $\hat{\mathbf{S}} = \mathcal{F}(n, L_s)$. 

\paragraph{Ambiguities in Shape from Polarization.} \label{sec:ambiguity}
Shape from Polarization suffers from intrinsic ambiguities.
AoLP is invariant under a $180^\circ$ rotation, leading to a $\pi$ ambiguity in azimuth.
Additionally, diffuse and specular reflections produce orthogonal polarization orientations, resulting in a $\pi/2$ ambiguity when the dominant reflection component is unknown.
These ambiguities motivate the integration of additional geometric priors using RGB cues, as we propose in our method.

\subsection{Monocular normal estimators}
\noindent A pretrained monocular estimator $f$ predicts the surface normal as
$
n = f(x),
$
by leveraging the rich visual priors learned from pretraining on large-scale datasets.
Deep-learning models span diverse architectures: iterative diffusion-based models (\eg, Marigold~\cite{marigold}), iterative flow-based models (\eg, Lotus-v2~\cite{He2025Lotus2AG}), feed-forward transformer models (\eg, MoGe-2~\cite{moge}).

\section{Method}
\label{sec:method}
\begin{figure}[ht]
  \centering
  \includegraphics[width=\textwidth]{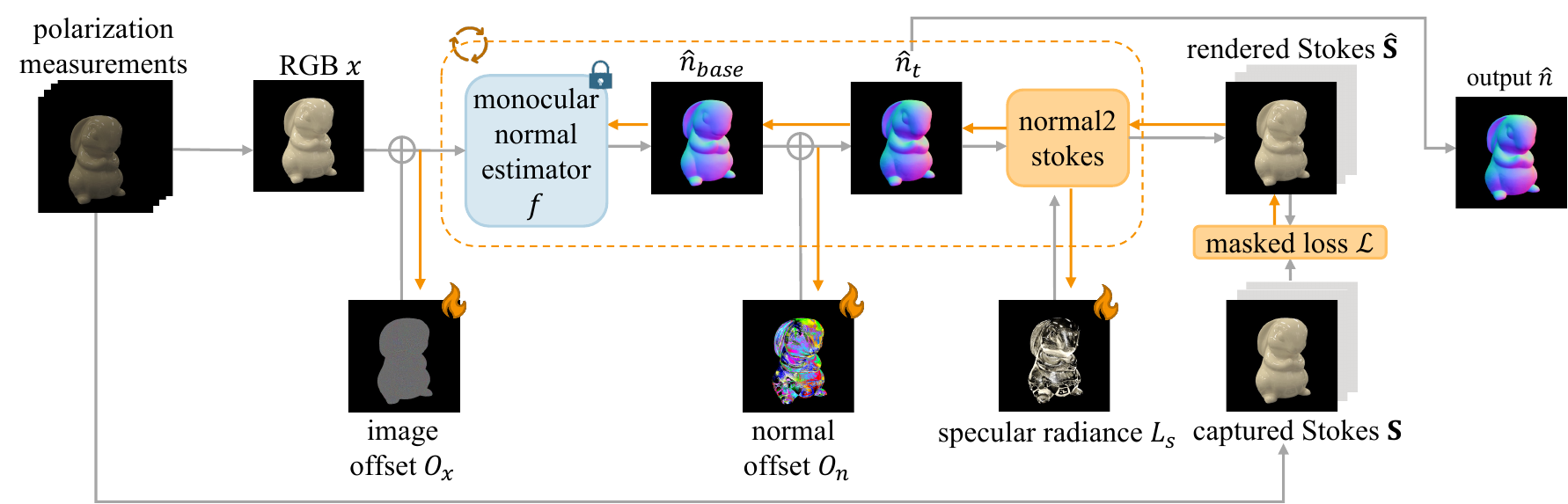}
  \caption{
    \textbf{Poppy pipeline.}
    Given polarization measurements, we compute the observed Stokes map $\mathbf{S}$, extract the RGB image $x$, and add learnable image offset $O_x$ to $x$. A frozen backbone produces base normals $\hat{n}_{\text{base}}$; a learnable normal offset $O_n$ yields the refined estimate $\hat{n}_t = \hat{n}_{\text{base}} + O_n$. Using Fresnel equations, the predicted Stokes $\hat{\mathbf{S}}$ is computed from $\hat{n}_t$ and specular radiance $L_s$. We minimize the polarization consistency loss between $\hat{\mathbf{S}}$ and $\mathbf{S}$ to update the image offset $O_x$, normal offset $O_n$, and specular map $L_s$ over $T$ steps, while keeping backbone weights fixed. 
  }
  \label{fig:pipeline}
\end{figure}
\noindent Given polarization measurements captured by a linear polarization camera, we recover surface normals that are consistent with physical light transport while preserving the strong geometric priors learned by modern RGB-based normal estimation networks. Instead of retraining a model, we formulate normal recovery as a test-time guidance problem. During the guidance, a differentiable rendering model enforces polarization constraints.

\subsection{Rendering Stokes from monocular normals}
\noindent Given the input polarization measurements $I_{\{0,45,90,135\}^{\circ}}$, we first compute the observed Stokes vector $\mathbf{S} = [S_0, S_1, S_2]^\top$ via Eq.~\ref{eq:stokes_linear}, where the unpolarized intensity $S_0$ serves as the RGB input $x$ to our pipeline.

A pretrained RGB normal estimator $f$ takes $x$ and produces a base surface normal estimate $\hat{n}_{\text{base}} = f(x)$. To synthesize the predicted Stokes vector $\hat{\mathbf{S}}$ from a given normal $\hat{n}_{\text{base}}$, we require the specular radiance $L_s$, following the procedure described in Sec.~\ref{sec:n2s}. In principle, $L_s$ can be optimized iteratively via polarization-guided refinement until the residual $|\mathbf{S} - \mathbf{\hat{S}}|$ vanishes. However, when $f$ produces an erroneous normal estimate in geometrically challenging regions, prolonged optimization of $L_s$ alone leads to either a persistent mismatch between the observed Stokes $\mathbf{S}$ and predicted Stokes $\mathbf{\hat{S}}=\mathcal{F}(f(x),L_s)$, or overfitting of $L_s$ to the incorrect normal (second column of Fig.~\ref{fig:image_behavior}(a))--- neither of which corrects the underlying geometric error.

This failure mode stems from the fact that the error originates in $\hat{n}_{\text{base}}$. To address this, we seek a test-time guidance mechanism that corrects the normal estimate directly, without modifying frozen model weights, and ensures compatibility with any pretrained monocular RGB normal estimator in a plug-and-play fashion, independent of its architecture.

\subsection{Polarization-guidance parameters}
\label{sec:guidance_params}

\begin{figure}
  \includegraphics[width=1.0\linewidth]{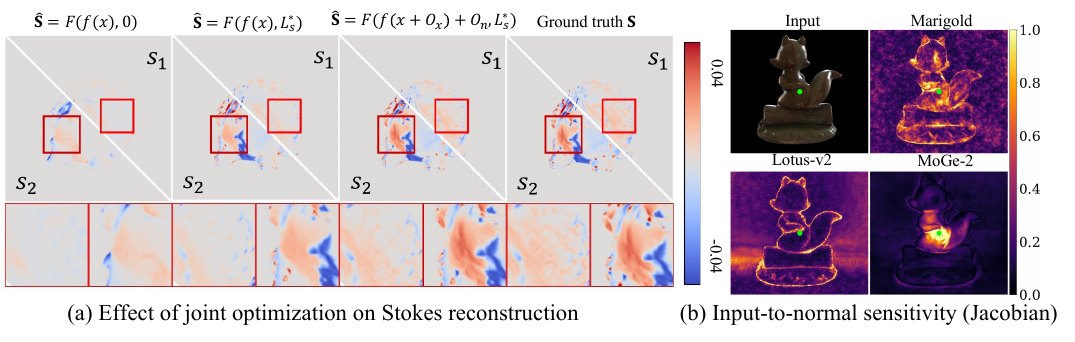}
\caption{
    (a)  Stokes reconstruction of hedgehog scene (from NeRSP) when:  $L_s=0$ (diffuse only) in column 1; only $L_s$ is learned from backbone predicted normals $\hat{n}_{\text{base}}$ in column 2; $L_s, O_{x},$ and $O_{n}$ are jointly learned in column 3.
    (b)  Jacobian magnitude maps for a selected input pixel, normalized by the 99th percentile for different backbones, showing how perturbations of a single input pixel (green dot) influence the output normal map at a global level, across spatial locations.}
  \label{fig:image_behavior}
\end{figure}

\noindent To correct the erroneous normal $\hat{n}_{\text{base}}$ while keeping the model weights frozen, we introduce learnable parameters applied before and after $f$. Formally, the refined normal at step $t$ is expressed as
$
\hat{n}_t = g_{\text{a}}(f(g_{\text{b}}(x))),
$
where $g_{\text{b}}$ and $g_{\text{a}}$ denote arbitrary functions applied to the input and output of $f$, respectively. We adopt the simplest instantiation, $g_{\text{b}}(x) = x + O_x$ and $g_{\text{a}}(\hat{n}_{\text{base}}) = \hat{n}_{\text{base}} + O_n$, yielding
$
\hat{n}_t = f(x + O_x) + O_n,
$
where $O_x \in \mathbb{R}^{H \times W \times 3}$ and $O_n \in \mathbb{R}^{H \times W \times 3}$ are per-pixel offsets applied to the input image $x$ and the predicted normal $\hat{n}_{\text{base}}$, respectively.
\paragraph{Global Guidance.}
The image offset $O_x$ is motivated by the observation that even a small perturbation in input space can produce a globally significant correction in the output normals. This behavior is explained by the Jacobian of different models $f$, visualized in Fig.~\ref{fig:image_behavior}(b), which shows that a change in a single input pixel has a broad influence over the output normal map (details in Sec.~\ref{sec:jacobian_computation} in the Supplement). Consequently, the image offset $O_x$ serves as an efficient mechanism for globally steering the predicted geometry toward a polarization-consistent solution.
\paragraph{Local Guidance.}
The normal offset $O_n$, applied directly to the output, addresses the tendency of monocular normal estimators to produce smooth predictions. The normal offset $O_n$ recovers high-frequency surface details that $f$ fails to capture. Due to its direct coupling with the normal output, the normal offset $O_n$ exhibits high sensitivity to the polarization loss. We therefore defer its optimization until $L_s$ has converged sufficiently to yield a reliable estimate of Stokes $\mathbf{\hat{S}}=\mathcal{F}(f(x + O_x) + O_n,L_s)$ (third column in Fig.~\ref{fig:image_behavior}(a)). We assume orthographic projection and ignore viewing directions, an approximation that introduces negligible error unless the object is captured at close range.
     
\subsection{Guidance objective and updates}
\label{sec:guidance_obj}
\noindent During test-time guidance, we convert the refined normal $\hat{n}_t$ into Stokes $\hat{\mathbf{S}}$ for polarization consistency check. We optimize the learnable parameters $\Theta = \{L_s, O_x, O_n\}$ iteratively over $T$ steps to minimize the discrepancy between the observed and predicted Stokes vectors. At each step $t$, the backbone $f$ takes the offset input $x + O_x$ and produces the base normal $\hat{n}_{\text{base}} = f(x + O_x)$. The normal offset $O_n$ is then applied to yield the refined normal $\hat{n}_t = \hat{n}_{\text{base}} + O_n$. Given $\hat{n}_t$ and the specular radiance $L_s$, the predicted Stokes vector $\mathbf{\hat{S}}(\hat{n}_t, L_s)$ is computed following Sec.~\ref{sec:n2s}. The polarization consistency loss is defined as:
\begin{equation}
    \mathcal{L} = \sum_p M(p) \sum_{i=0}^{2} \left| S_i(p) - \widehat{S}_i(p) \right|,
\end{equation}
where $p$ denotes a pixel and $M(p)$ is a validity mask that restricts the loss to physically meaningful measurements. Specifically, a pixel is included if it has sufficient signal ($S_0 > 0.01$), is not saturated ($S_0 < 1$), and satisfies the physical Stokes constraint ($S_1^2 + S_2^2 \leq S_0^2$).

\paragraph{Global-then-local guidance.}The three parameters are optimized in a staged schedule, each with a distinct learning rate. In the early phase ($t < 50$), we only update $L_s$ and $O_x$, allowing the specular radiance to converge and the predicted normals to reach global geometric plausibility. At $t = 50$, $O_n$ is introduced to refine high-frequency surface details that the backbone fails to capture, and all three parameters are subsequently updated until $t = T$.

The learnable parameters 
$\Theta \in \{L_s,O_x,O_n\}$ are updated with an optimizer
using gradient descent
$\Theta^{(k+1)}
=
\Theta^{(k)} - \lambda_\Theta \nabla_\Theta \mathcal{L}$.
After $T$ iterations, the output surface normal is
$
\hat{n}
= 
\frac{f(x+O_x) + O_n}
{\|f(x+O_x) + O_n\|_2}.
$

The learned specular radiance $L_s$ and the refined output normal $\hat{n}$ together synthesize the predicted Stokes vector $\hat{\mathbf{S}}$ that closely matches the observed Stokes vector $\mathbf{S}$ as shown in the third column in Fig.~\ref{fig:image_behavior}(a). Consequently, minimizing this discrepancy drives the predicted normal $\hat{n}$ toward the true surface normal.

\section{Results}
\label{sec:experiments}

\begin{table*}[t]
  \centering
\caption{\textbf{Aggregated normal estimation performance on real and synthetic
  benchmarks.} Our polarization guidance (+ Poppy) consistently reduces MAE and RMSE across backbone RGB estimators while improving
  thresholded accuracy, confirming gains in both global surface orientation and fine-scale local geometric
  detail. +Poppy denotes joint optimization of the image offset and normal offset. Best result per backbone within each data split is highlighted in bold.}
  \label{tab:full_metrics}
  \resizebox{\linewidth}{!}{
    \begin{tabular}{lcccccccccccc}
      \toprule
      & \multicolumn{6}{c}{\textbf{Real}}
      & \multicolumn{6}{c}{\textbf{Synthetic}} \\
      \cmidrule(lr){2-7} \cmidrule(lr){8-13}
      Method
      & Mean & Median & RMSE & Acc11.25 & Acc22.5 & Acc30
      & Mean & Median & RMSE & Acc11.25 & Acc22.5 & Acc30 \\
      \midrule

      SfPUEL\cite{sfpuel}
      & 20.68 & 17.08 & 26.15 & 0.37 & 0.70 & 0.80
      & 17.03 & 12.91 & 23.33 & 0.46 & 0.79 & 0.87 \\

      DSINE\cite{dsine}
      & 23.06 & 19.07 & 28.83 & 0.27 & 0.62 & 0.76
      & 24.13 & 19.94 & 30.00 & 0.26 & 0.60 & 0.75 \\

      Lotus\cite{lotus}
      & 16.69 & 13.96 & 21.70 & 0.42 & 0.78 & 0.88
      & 19.99 & 17.04 & 25.06 & 0.30 & 0.69 & 0.84 \\

      StableNormal\cite{stablenormal}
      & 17.80 & 14.77 & 23.29 & 0.41 & 0.75 & 0.86
      & 18.45 & 14.65 & 24.55 & 0.39 & 0.75 & 0.86 \\

      \midrule

      Marigold\cite{marigold}
      & 18.18 & 15.25 & 23.30 & 0.36 & 0.74 & 0.86
      & 20.99 & 17.40 & 26.43 & 0.30 & 0.68 & 0.82 \\

      Marigold + Poppy
      & \textbf{15.28} & \textbf{12.57} & \textbf{20.26}
      & \textbf{0.47} & \textbf{0.83} & \textbf{0.92}
      & \textbf{15.60} & \textbf{11.89} & \textbf{22.08}
      & \textbf{0.56} & \textbf{0.82} & \textbf{0.89} \\

      \midrule

      Lotus-v2\cite{He2025Lotus2AG}
      & 14.68 & 12.05 & 19.51 & 0.49 & 0.85 & 0.93
      & 16.52 & 13.44 & 22.08 & 0.42 & 0.81 & 0.90 \\

      Lotus-v2 + Poppy
      & \textbf{12.65} & \textbf{10.05} & \textbf{17.62}
      & \textbf{0.61} & \textbf{0.89} & \textbf{0.94}
      & \textbf{12.26} & \textbf{8.81} & \textbf{18.70}
      & \textbf{0.66} & \textbf{0.89} & \textbf{0.93} \\

      \midrule

      MoGe-2\cite{moge}
      & 13.10 & 10.51 & 18.11 & 0.57 & 0.87 & 0.94
      & 14.13 & 11.33 & 20.01 & 0.51 & 0.87 & 0.94 \\

      MoGe-2 + Poppy
      & \textbf{12.26} & \textbf{9.88} & \textbf{16.99}
      & \textbf{0.61} & \textbf{0.90} & \textbf{0.95}
      & \textbf{10.89} & \textbf{8.05} & \textbf{16.41}
      & \textbf{0.70} & \textbf{0.92} & \textbf{0.96} \\

      \bottomrule
    \end{tabular}
  }
\end{table*}

\subsection{Implementation details}
\label{sec:impl_details}

\paragraph{Datasets.}
Synthetic benchmarks include SfPUEL~\cite{sfpuel} (low SNR, highly specular), NeRSP~\cite{nersp} (high reflectance, low illumination), NeISF~\cite{neisf} (moderate reflectance), and DeepPol~\cite{deeppol} (mixed diffuse and specular).
Real-world benchmarks include SfPUEL~\cite{sfpuel}, NeRSP~\cite{nersp}, and PISR~\cite{pisr} (highly reflective, textureless), with ground-truth normals extracted from provided 3D meshes.

\paragraph{Baseline models.}
We compare against the polarization-based model SfPUEL~\cite{sfpuel} and RGB-based methods DSINE~\cite{dsine}, StableNormal~\cite{stablenormal}, Marigold~\cite{marigold}, Lotus-v2~\cite{He2025Lotus2AG}, and MoGe-2~\cite{moge}, all under their default configurations.
Poppy is applied to Marigold, Lotus-v2, and MoGe-2, representing diffusion-based, flow-based, and feed-forward backbones.
We report mean angular error (MAE), median angular error (Median), angular RMSE, and accuracy under angular thresholds.

\paragraph{Optimization.}
We optimize with Adam for 100 steps. Learning rates: $\lambda_{L_s} = 0.01$; backbone-dependent rates for $O_x$: $10^{-4}$ (Marigold), $5\times10^{-4}$ (Lotus-v2), $10^{-5}$ (MoGe-2). The normal offset $O_n$ uses $\lambda_{n} = 0.001$ across all backbones and activates at step 50, once $L_s$ has converged. 

\paragraph{Backbone inference.} For Marigold, we use 4 denoising steps with 25 guidance steps each, without ensemble inference. For Lotus-v2, guidance precedes the detail-sharpening stage, followed by 10 sharpening iterations at default settings. For MoGe-2, gradients are backpropagated directly through the feed-forward pipeline. See Sec.~\ref{sec:backbone_details} in the Supplement for per-backbone details.

\paragraph{Runtime and memory.} At $768\times768$ resolution, per-step cost splits roughly 45/55 between inference and optimization. Per-step costs: 567~ms (MoGe-2), 950~ms (Marigold), 1729~ms (Lotus-v2), yielding total runtimes of 57~s, 99~s, and 173~s at 100 steps.
Peak memory: 12.97~GB (MoGe-2), 35.13~GB (Marigold), 66.64~GB (Lotus-v2), dominated by gradient retention during optimization.
All experiments run on NVIDIA RTX PRO 6000 Blackwell GPUs (95.6~GB VRAM).

\begin{figure}[t]
    \centering
    \includegraphics[width=1\linewidth]{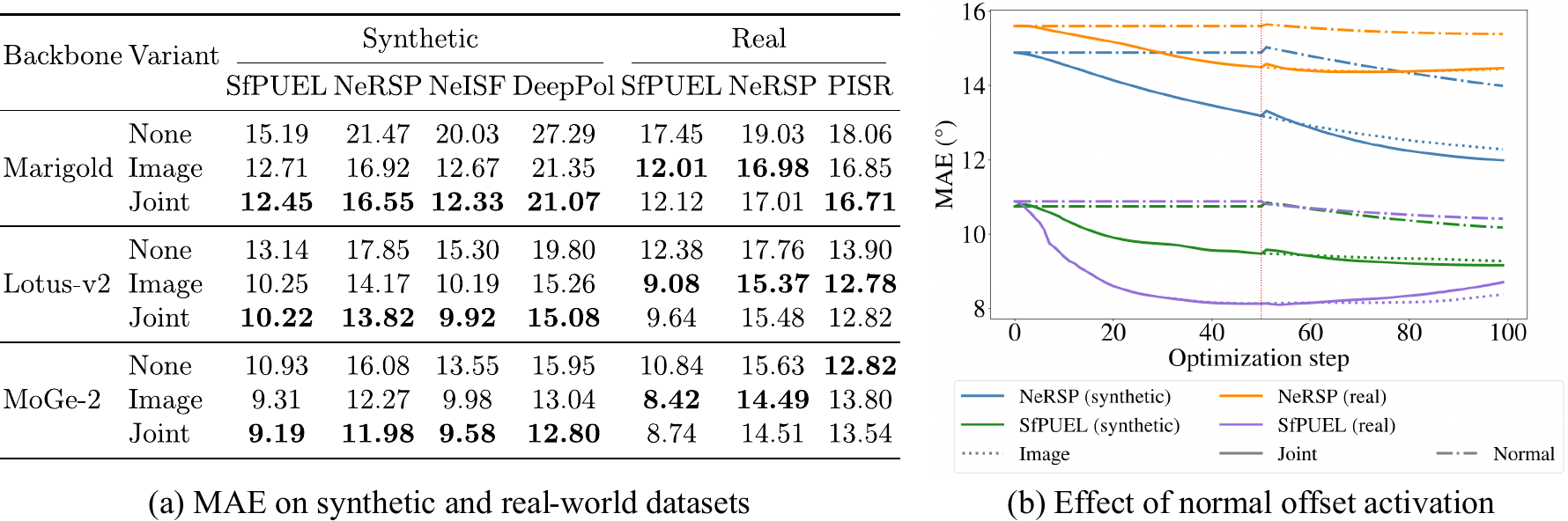}
\caption{
  \textbf{Ablation of guidance variants on real and synthetic datasets.}
  (a) Mean angular error (MAE) across three backbones (Marigold, Lotus-v2, MoGe-2) with no guidance (None), image offset guidance (Image), and joint image and normal offsets guidance (Joint). Guidance improves normals, with joint guidance performing slightly better on synthetic datasets.
  (b) MAE over optimization steps on SfPUEL and NeRSP with MoGe-2 backbone.
  The red-dashed line at $t{=}50$ marks the activation of the normal offset.
  On synthetic data, joint guidance accelerates error reduction.
  On real data, sensor noise causes the normal offset to slightly increase MAE, though high-frequency detail is still recovered.
}
    \label{fig:normal_offset}
\end{figure}
\begin{figure}[t]
  \centering
  \includegraphics[width=1.0\linewidth]{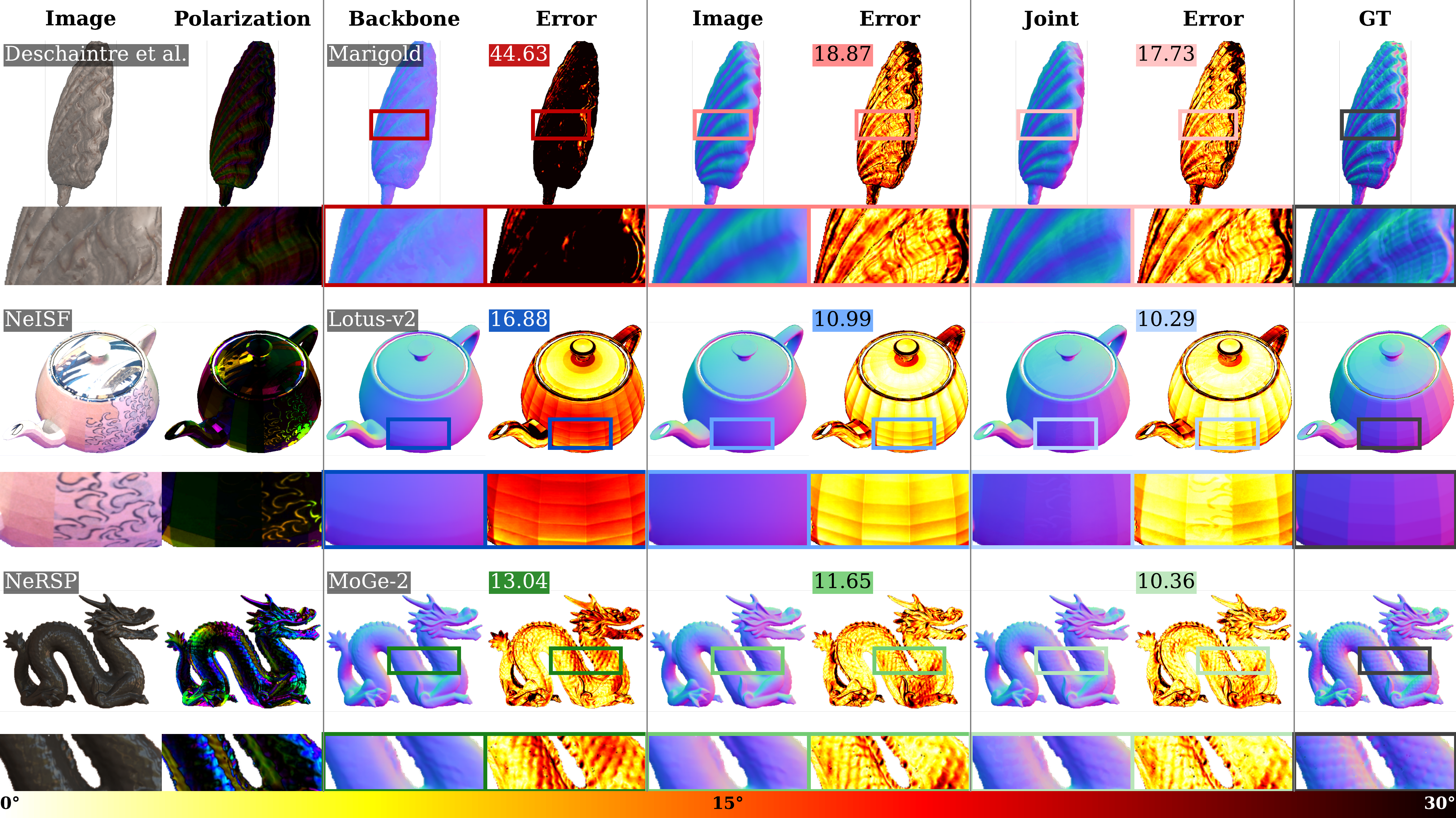}
  \caption{
  \textbf{Comparison of image (global) and joint (global-then-local) guidance on synthetic data.}
We compare output normals under three guidance conditions across three backbones: no guidance (Backbone), image offsets (Image), and joint image and normal offsets (Joint).
  Joint guidance recovers fine geometric structures -- shell ridges, mesh facets, and dragon scales -- that image-offset guidance alone cannot resolve.
  Pixel-wise angular error maps (brighter is lower error) and mean angular error are also shown for each prediction.
  Best viewed at high magnification.
}
  \label{fig:offset_ablation}
\end{figure}

\begin{figure}[t]
  \centering
  \includegraphics[width=1\linewidth]{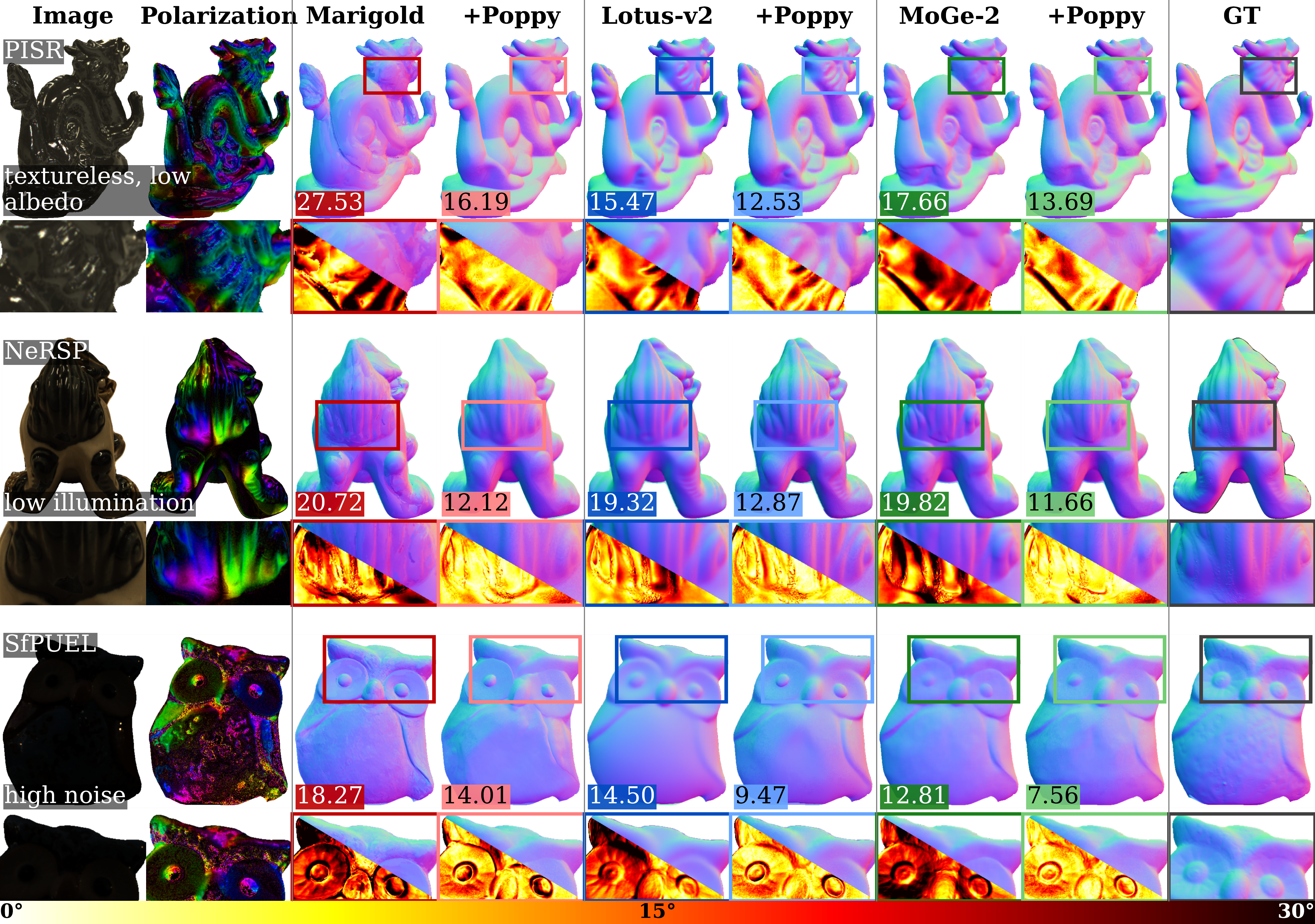}
  \caption{
  \textbf{Qualitative comparison on real-world scenes.}
  Columns show backbone predictions and results after Poppy guidance.
  Pixel-wise angular error maps accompany each prediction (brighter is lower error), with the mean angular error annotated. Scenes span three challenging categories: textureless and highly reflective objects, and low-SNR conditions (low illumination, low albedo, high noise). Poppy shows consistent improvement in normal quality across all categories and backbones.
}
  \label{fig:baseline_ablation}
\end{figure}

\subsection{Results and analysis}
\paragraph{Quantitative comparisons with monocular normal baselines.}

Tab.~\ref{tab:full_metrics} reports aggregated performance across all backbones and benchmarks. Adding polarization guidance (+Poppy) consistently improves MAE on both real and synthetic datasets, with MAE reductions of 6--26\% and RMSE reductions of 7--18\%, reflecting improved global surface orientation consistency. More notably, Acc11.25 (the fraction of pixels with angular error below 11.25$^\circ$) improves by 7--31\% on real data and 37--87\% on synthetic data, demonstrating that Poppy not only corrects large-scale geometric errors but also recovers high-frequency surface details that RGB-based backbones systematically fail to capture.

\paragraph{Qualitative results on real-world scenes.}
To illustrate how these aggregate gains manifest in individual scenes, Fig.~\ref{fig:baseline_ablation} presents qualitative comparisons on real-world scenes that are particularly challenging for RGB-based estimators. These scenes include textureless low-albedo surfaces, highly specular objects, and low-SNR conditions arising from low illumination, low albedo, or high measurement noise. Across all three backbone architectures and all challenging conditions, Poppy yields MAE reductions of 19--41\% over the backbone predictions, with the largest improvements on scenes where RGB-based appearance cues alone are most ambiguous.

\paragraph{Effect of global and local guidance.}
Fig.~\ref{fig:normal_offset} and Fig.~\ref{fig:offset_ablation} analyze the individual contributions of the image offset $O_x$ and the normal offset $O_n$. In Fig.~\ref{fig:normal_offset}(a), the image offset $O_x$ alone accounts for the majority of the MAE reduction across all backbones and datasets, correcting large-scale geometric errors toward a globally polarization-consistent solution. Adding the normal offset $O_n$ (Joint) yields a further reduction on synthetic data, though the additional gain is modest compared to the image offset alone. Across all backbones and datasets, Poppy outperforms the backbone baseline, with the sole exception of MoGe-2 on PISR, which we attribute to pronounced perspective distortion that our orthographic assumption does not model. See Sec.~\ref{sec:fov} in the Supplement for perspective-aware guidance by incorporating field-of-view.

The per-step MAE curve in Fig.~\ref{fig:normal_offset}(b) is consistent with this global-then-local behavior: MAE decreases primarily during the early optimization phase under the image offset; at $t{=}50$, when the normal offset is activated on synthetic data, a further sharp drop occurs as joint offset learning refines the remaining errors. Once jointly optimized (Fig.~\ref{fig:offset_ablation}), the normal offset recovers fine-scale geometric structures that the backbone over-smooths, such as shell ridges, mesh facets, and dragon scales (zoomed insets). 

\begin{figure}[t]
  \centering
  \includegraphics[width=1.0\linewidth]{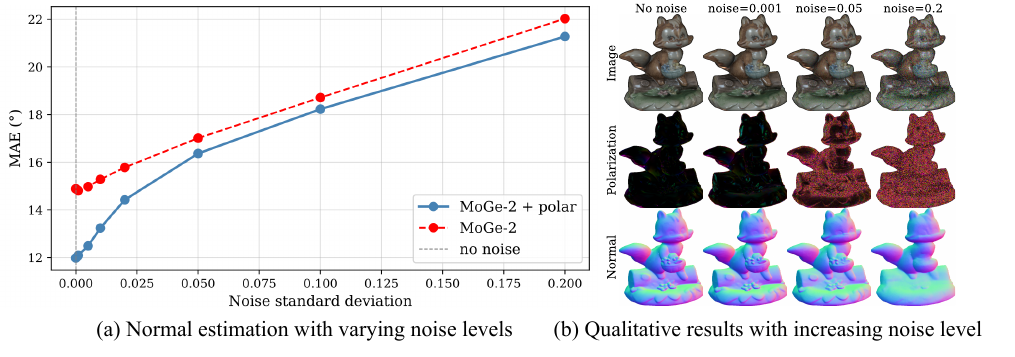}
  \caption{
    \textbf{Sensitivity to input noise.}
  We add zero-mean Gaussian noise of increasing standard deviation $\sigma$ to the NeRSP synthetic measurements.
  The plot reports mean angular error (MAE) for MoGe-2 with and without polarization guidance. Guidance improves accuracy at low noise levels, but the advantage diminishes with increasing $\sigma$.
  }
  \label{fig:noise_vs_mae}
\end{figure}

\paragraph{Robustness to sensor noise.}
 We examine how Poppy behaves when the polarization input itself is corrupted. Fig.~\ref{fig:noise_vs_mae} reports MAE after adding zero-mean Gaussian noise of varying standard deviation $\sigma$ to the NeRSP synthetic polarization measurements. At low $\sigma$, Poppy reduces MAE relative to the no-guidance prediction on MoGe-2. As $\sigma$ grows, the guided result converges toward the unguided backbone prediction -- noisy cues are effectively ignored rather than amplified. 

\begin{figure}[h!]
  \centering
  \includegraphics[width=1.0\linewidth]{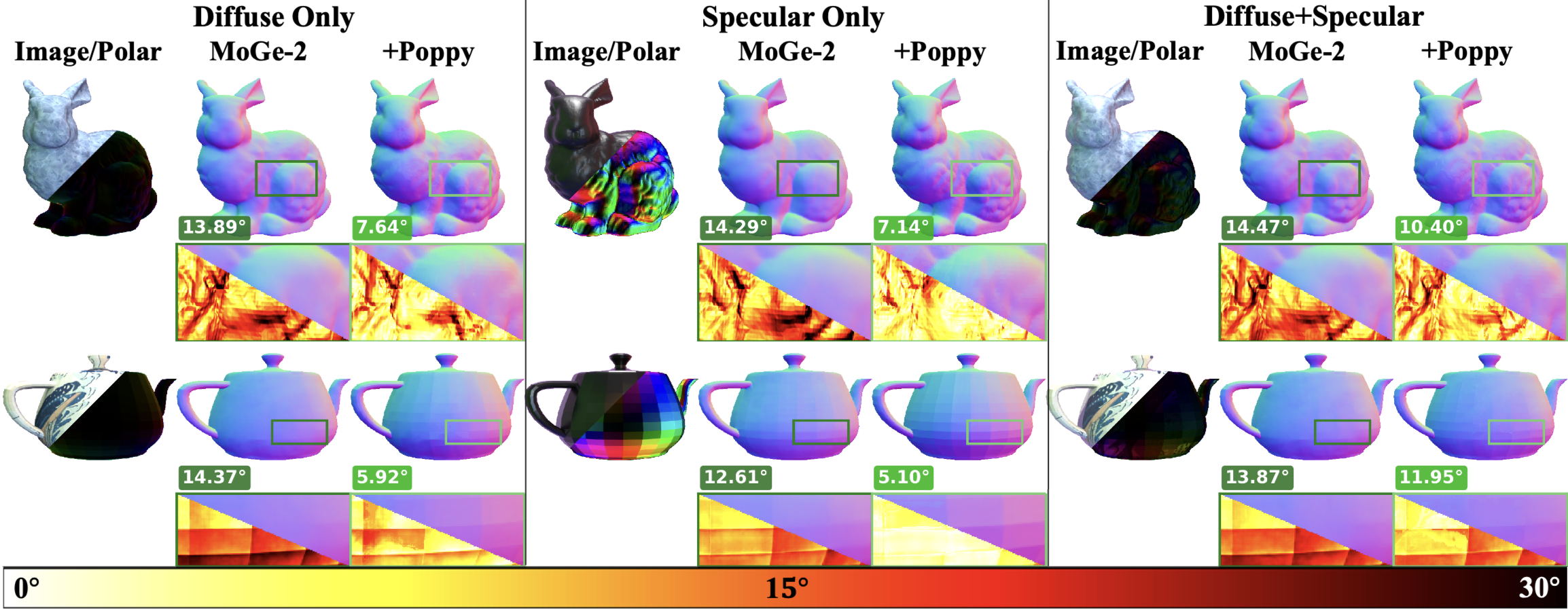}
\caption{\textbf{Robustness to material reflectance properties.}                
    Using the NeISF dataset, we render scenes with purely specular, purely
  diffuse, and mixed diffuse and specular reflectance to evaluate polarization guidance under varying material properties. Specular reflectance produces the strongest polarization signal and yields the largest improvement, followed by diffuse reflectance. Mixed reflectance case yields smaller gains due to the challenge of accurately decomposing diffuse and specular components. It shows pixel-wise angular error maps (brighter is lower error) with the mean angular
  error annotated.
  }
  \label{fig:diff_spec_diff+spec}
\end{figure}

\paragraph{Robustness to material properties.}
We isolate the effect of material reflectance using the NeISF synthetic dataset (Fig.~\ref{fig:diff_spec_diff+spec}), testing three conditions: purely diffuse, purely specular, and mixed reflectance. MoGe-2 without guidance achieves comparable MAE across all three conditions, but polarization guidance (+ Poppy) shows the most refinement on the specular case, where specular reflection produces stronger linear polarization. Diffuse surfaces yield the second-largest improvement, as even weaker polarization from diffuse radiance carries useful orientation information. In the mixed case, accurate diffuse-specular separation becomes a prerequisite; errors in this decomposition propagate into the polarization guidance, resulting in smaller but still meaningful gains.

\subsection{Applications}
\label{sec:applications}
\paragraph{Appearance decomposition.}
Beyond refining normals, Poppy yields a physics-based decomposition of captured appearance ($S_0$) into diffuse and specular components ($L_d$ and $L_s$), as shown in Fig.~\ref{fig:radiance_decomposition} (top row). We provide image editing applications in Sec.~\ref{sec:image_editing} in the Supplement. From the decomposed appearance and refined normals, our forward model synthesizes the polarization properties of diffuse and specular components (Fig.~\ref{fig:radiance_decomposition} bottom row).

\begin{figure}[t]
  \centering
  \includegraphics[width=1.0\linewidth]{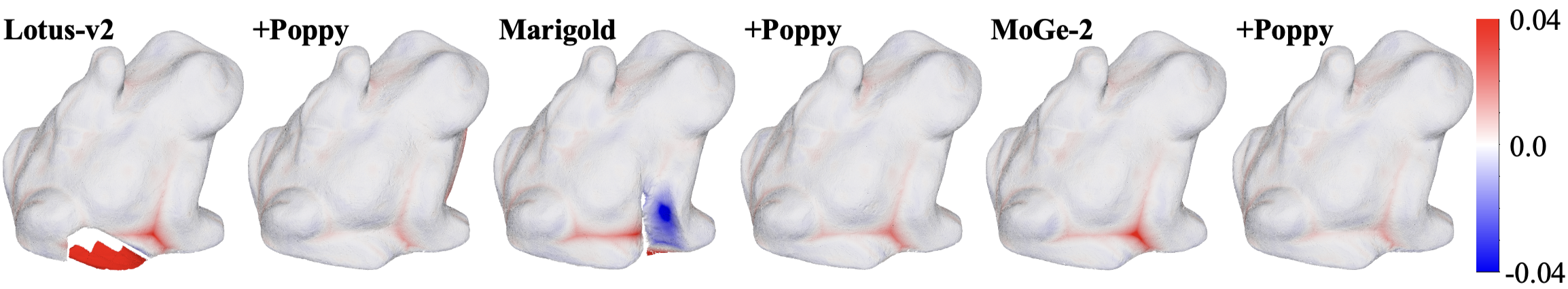}
  \caption{\textbf{Mesh reconstruction with refined surface normals.} 
    We reconstruct meshes using VCR-GauS~\cite{chen2024vcr} from multi-view images and corresponding normals estimated with and without polarization guidance.
    Polarization-refined normals yield better mesh quality across the three backbones. Signed distance to the ground-truth mesh is shown for each reconstruction.
  }

  \label{fig:mesh_reconstruction}
\end{figure}

\paragraph{3D mesh reconstruction.}
In Fig.~\ref{fig:mesh_reconstruction}, we test our improved normals on downstream mesh reconstruction tasks. We use VCR-GauS~\cite{chen2024vcr} as our baseline to construct surface mesh, guided by multi-view RGB images and monocular surface normal estimates on those views. We observe an improvement in reconstructed mesh accuracy with our surface normals and an $\approx$6\% improvement in Chamfer distance as compared to RGB-only backbones. Note that we attempted using Gaussian Surfels~\cite{Dai2024GaussianSurfels}. However, it failed due to poor initialization points retrieved by COLMAP~\cite{schoenberger2016sfm, schoenberger2016mvs} on glossy, textureless surfaces. 

\section{Conclusion}
\label{sec:conclusion}
\noindent We introduced Poppy, a training-free framework that guides frozen RGB normal estimators with physics-based polarization constraints at test time, refining normals on reflective, textureless, and dark surfaces without polarimetric training data or backbone retraining.
A learnable image offset propagates global corrections through the backbone; a normal offset recovers high-frequency local surface detail; and a learned specular radiance map resolves diffuse--specular ambiguity in single-view polarization.
Because guidance parameters lie entirely outside the backbone, Poppy applies to diffusion-based~\cite{marigold}, flow-based~\cite{He2025Lotus2AG}, and feed-forward~\cite{moge} backbones in a plug-and-play manner.
Across seven benchmarks, the framework reduces mean angular error by 23--26\% on synthetic data and 6--16\% on real captures.

Although we focus on showing that polarization can guide any frozen monocular estimator at test time without retraining, our framework opens several directions for future work.
The iterative optimization requires multiple forward passes through the backbone; however, we expect that better optimization strategies can reduce the computational overhead.
The learnable image and normal offsets are lightweight yet effective for normal refinement; replacing them with neural network adapters could further increase the representation capacity at the expense of requiring minimal polarization data for fine-tuning.
Extending robustness to higher sensor noise levels, handling perspective distortions for close-range captures, and modeling more complex material types such as metallic, transparent, and translucent surfaces are all within reach of the modular framework.
We see this work as a step toward steering frozen foundation models with physics-based cues, available solely at test time.

\bibliographystyle{splncs04}
\bibliography{main}

\newpage
\appendix
\section{Visualizing input-to-normal sensitivity using Jacobians}
\label{sec:jacobian_computation}
\noindent Here we describe the visualization used in Fig.~\ref{fig:image_behavior}(b) and Sec.~\ref{sec:guidance_params} to characterize input-to-normal sensitivity of monocular normal estimators.
The model $f$ maps an input RGB image 
$I \in \mathbb{R}^{3 \times H \times W}$
to a surface normal map
$n \in \mathbb{R}^{H \times W \times 3},
$
where the output channels correspond to the normal components
$n[y,x] = (n_x, n_y, n_z).
$

We analyze how perturbing a single input pixel affects the predicted normals across the entire output image.  
Let the chosen input pixel location be
$(p_i, p_j).
$
Our goal is to measure how sensitive each output pixel $(i',j')$ is to small perturbations of this input pixel.

\paragraph{Step 1: Computing Jacobian–Vector Products.}

The full Jacobian of the network is
$J_f(I) = \frac{\partial n}{\partial I},$
which maps perturbations in the input image to perturbations in the output normals.
For each input color channel \(c_k \in \{R,G,B\}\), we construct a tangent vector
$v^{(c_k)} \in \mathbb{R}^{3 \times H \times W},$
defined as
\[
v^{(c_k)}_{i,j,c} =
\begin{cases}
1 & \text{if }  i = p_i,\; j = p_j, \; c = c_k\\
0 & \text{otherwise}.
\end{cases}
\]

This vector represents an infinitesimal perturbation of a single input variable \(I[c_k,p_i,p_j]\).
Using forward-mode automatic differentiation, we compute the Jacobian–vector product
$\mathrm{JVP}^{(c_k)} = J_f(I)\, v^{(c_k)}.$
Because \(v^{(c_k)}\) is a one-hot vector in the input space, the JVP extracts exactly one column of the full Jacobian:

\[
\mathrm{JVP}^{(c_k)}[c_o, i',j']
=
\frac{\partial n[c_o,i',j']}{\partial I[c_k, p_i,p_j]},
\]

where \(c_o \in \{n_x,n_y,n_z\}\) denotes the output normal component.
Repeating this procedure for the three input channels provides all partial derivatives relating the selected input pixel to the output normals.

\paragraph{Step 2: Local Jacobian Block.}

For a fixed output pixel \((i',j')\), we assemble a \(3\times3\) local Jacobian block

\[
J_{i',j'} =
\begin{pmatrix}
\dfrac{\partial n_x[i',j']}{\partial I[R,p_i,p_j]} &
\dfrac{\partial n_x[i',j']}{\partial I[G,p_i,p_j]} &
\dfrac{\partial n_x[i',j']}{\partial I[B,p_i,p_j]}
\\[8pt]
\dfrac{\partial n_y[i',j']}{\partial I[R,p_i,p_j]} &
\dfrac{\partial n_y[i',j']}{\partial I[G,p_i,p_j]} &
\dfrac{\partial n_y[i',j']}{\partial I[B,p_i,p_j]}
\\[8pt]
\dfrac{\partial n_z[i',j']}{\partial I[R,p_i,p_j]} &
\dfrac{\partial n_z[i',j']}{\partial I[G,p_i,p_j]} &
\dfrac{\partial n_z[i',j']}{\partial I[B,p_i,p_j]}
\end{pmatrix}.
\]

Each column of this matrix corresponds to the JVP result for a particular input channel.

\paragraph{Step 3: Frobenius Norm of the Local Jacobian.}

To summarize the total sensitivity of output pixel \((i',j')\) to perturbations of the selected input pixel, we compute the Frobenius norm of this local Jacobian block:

\[
\|J_{i',j'}\|_F
=
\sqrt{
\sum_{c_k \in \{R,G,B\}}
\sum_{c_o \in \{n_x,n_y,n_z\}}
\left(
\frac{\partial n[c_o,i',j']}
{\partial I[c_k,p_i,p_j]}
\right)^2
}.
\]

\paragraph{Interpretation.}

The value \(\|J_{y,x}\|_F\) measures the overall sensitivity of the predicted normal at pixel \((i',j')\) to perturbations of the input pixel \((p_i,p_j)\), aggregated across all input and output channels. Pixels that are weakly influenced by the selected input location will have
$\|J_{y,x}\|_F \approx 0,$
while pixels strongly coupled through the network will exhibit larger values. This Jacobian norm therefore provides a spatial map describing how perturbations at a single input pixel propagate through the model to affect the predicted surface normals globally as described in Sec.~\ref{sec:guidance_params}. The Jacobian magnitude map for a single-pixel perturbation is shown in Fig.~\ref{fig:image_behavior}(b).

\section{Effect of image guidance on the network input}
\begin{figure}
    \centering
    \includegraphics[width=.48\linewidth]{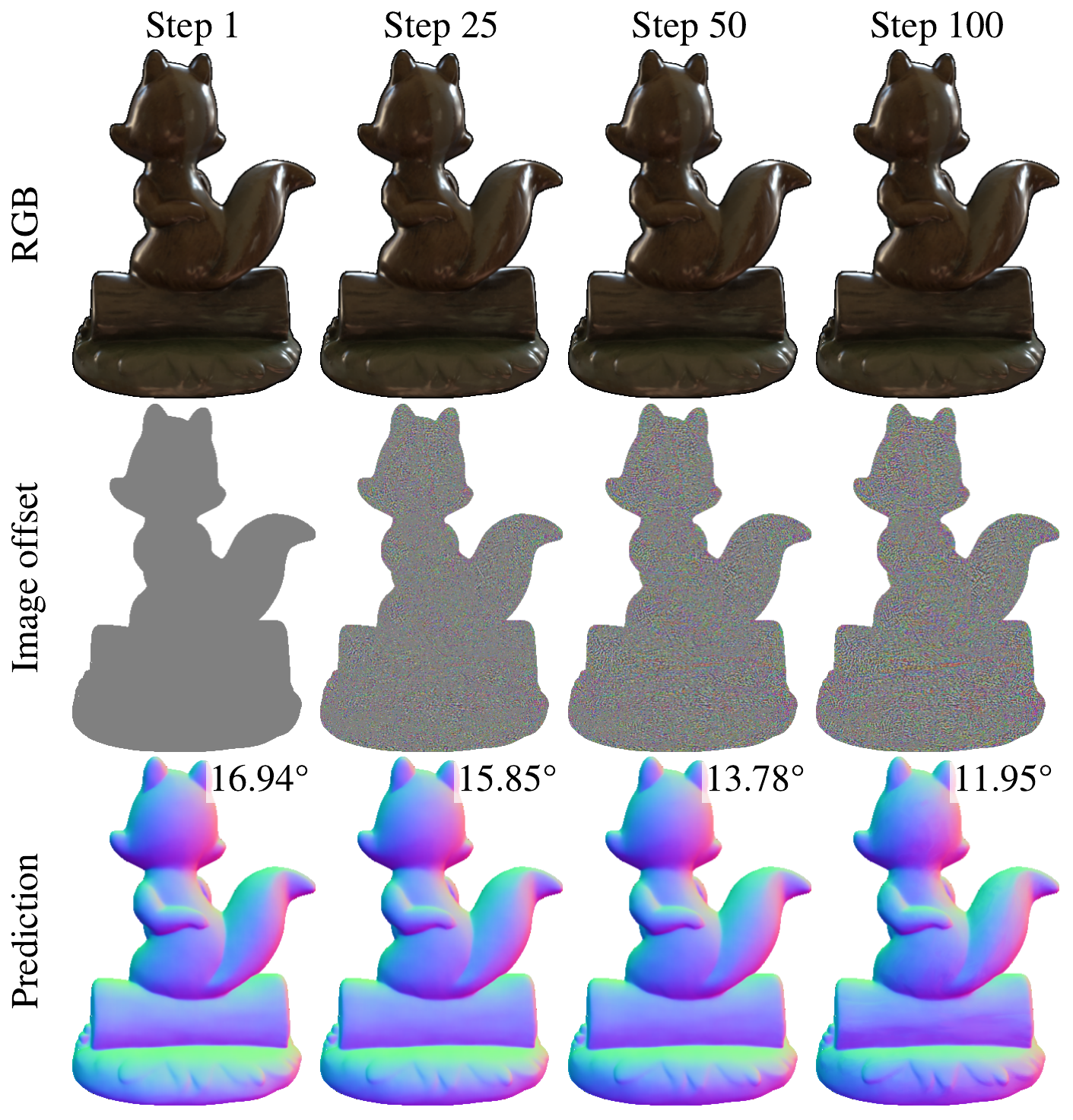}
\caption{
\textbf{Test-time image optimization behavior.}
The input image is iteratively refined using a learned offset.
The image offset is multiplied by 50 for visualization. Although the perturbations appear as nearly imperceptible noise in the RGB image, it produces large changes in the predicted surface normals.
The optimization progressively improves the normal prediction, reducing the mean angular error from $16.94^\circ$ to $11.95^\circ$.
}
    \label{fig:image_variation}
\end{figure}
\noindent Fig. ~\ref{fig:image_variation} illustrates the effect of the proposed test-time image optimization.
Although the image offset changes the input image only slightly during optimization, the predicted surface normals show substantial improvement.
The image offset appears as small high-frequency noise in the image space and is nearly imperceptible to human eyes.
Nevertheless, these small perturbations significantly change the network output, reducing the mean angular error from $16.94^\circ$ to $11.95^\circ$.

This phenomenon resembles adversarial perturbations~\cite{adversarial} in deep networks.
In adversarial examples, carefully designed small perturbations can dramatically alter the network prediction while remaining visually imperceptible.
In our setting, however, the perturbation is not used to degrade the model but instead to guide the network toward a physically consistent solution.
The image offset effectively acts as small perturbations that exploit the model's sensitivity to correct geometric errors while preserving the visual appearance of the input image.

\section{Backbone inference details}
\label{sec:backbone_details}
\begin{figure}[h!]
    \centering
    \includegraphics[width=1\linewidth]{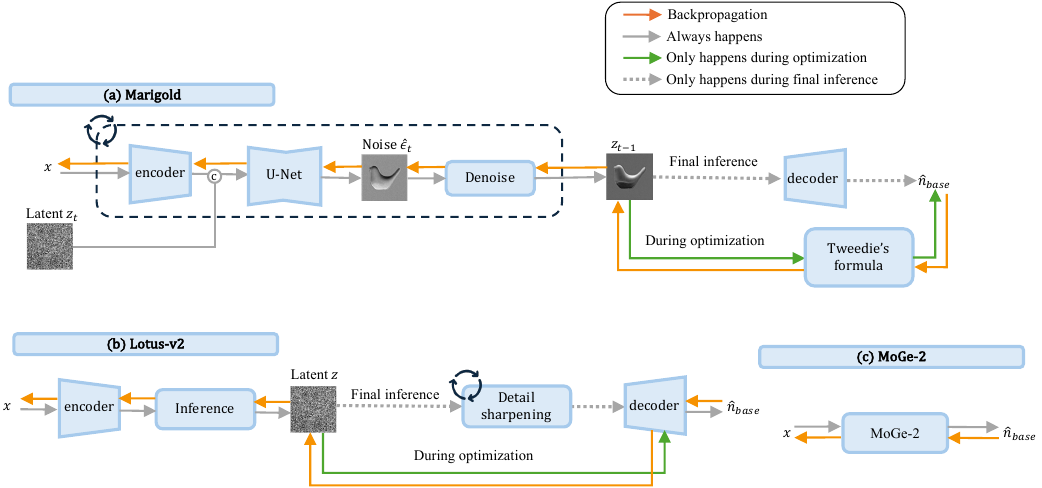}
\caption{
\textbf{Optimization details for different backbone architectures in Poppy.}
(a) Marigold performs optimization at each denoising step using the normal preview obtained via Tweedie’s formula.
(b) Lotus-v2 performs optimization during the latent inference stage before the detail sharpening step.
(c) MoGe-2 is a feed-forward network, allowing gradients to directly propagate to the input image during optimization.
}    \label{fig:implem_details}
\end{figure}
\noindent The proposed Poppy optimization framework is compatible with different types of monocular normal estimation models.
In Sec.~\ref{sec:impl_details}, we evaluate three representative backbones with distinct inference pipelines: a diffusion-based model (Marigold~\cite{marigold}), a multi-stage flow-based model (Lotus-v2~\cite{He2025Lotus2AG}), and a feed-forward model (MoGe-2~\cite{moge}), as illustrated in Fig.~\ref{fig:implem_details}.

\paragraph{Marigold} predicts surface normals through an iterative denoising process.
As shown in Fig.~\ref{fig:implem_details}(a), multiple optimization steps are performed at each denoising step.
At every DDIM~\cite{ddim} step, a normal preview is computed using Tweedie’s formula~\cite{Efron2011TweediesFA} from the predicted noise, and the polarization loss is evaluated on this preview.
Rather than backpropagating through the entire diffusion trajectory, optimization is applied locally at each denoising step.
This significantly reduces memory consumption and allows Poppy to accommodate arbitrary numbers of denoising steps selected at inference time.

\paragraph{Lotus-v2} follows a multi-stage inference pipeline consisting of latent inference, latent refinement (detail sharpening), and decoding, as illustrated in Fig.~\ref{fig:implem_details}(b).
Poppy optimization is applied during the latent inference stage.
The model first predicts a latent representation from the input image, which is then decoded to obtain the surface normal prediction used to compute the polarization loss.

Although the subsequent sharpening stage refines the latent representation, it operates only on the latent variables and does not directly depend on the input image.
Therefore, the optimization focuses on the latent inference stage, where the image influences the prediction.
As in the Marigold case, we avoid backpropagating through the entire inference pipeline because it would be memory-inefficient, especially when multiple sharpening steps are used.
Instead, the optimization updates the image based on the latent prediction produced during the initial inference step.
After optimization, the resulting latent representation continues through the remaining refinement and decoding stages during inference.
This design allows Poppy to remain memory-efficient while accommodating an arbitrary number of sharpening steps.

\paragraph{MoGe-2} is a feed-forward monocular normal estimator.
As shown in Fig.~\ref{fig:implem_details}(c), the network directly predicts normals from the input image in a single forward pass.
Thus, the polarization loss can be backpropagated directly to the image without any intermediate inference stages.
This makes Poppy straightforward for feed-forward architectures.

Across all architectures, the backbone networks remain frozen, and Poppy optimizes the learnable offsets before and after the network using polarization-guided losses (Sec.~\ref{sec:guidance_obj}).
The optimization is inserted at architecture-specific points in the inference pipeline, enabling a unified framework compatible with diffusion-based, multi-stage flow-based, and feed-forward normal estimation models.

\section{Incorporating field-of-view estimation from MoGe}
\label{sec:fov}
\begin{table}[h!]
\centering
\caption{
\textbf{Quantitative comparison on the PISR dataset.}
We compare the backbone model (MoGe-2), MoGe-2 with our polarization guidance assuming orthogonal projection (+ Poppy), and our guidance with perspective projection when the field-of-view (FoV) is incorporated (+ Poppy (FoV)).
In the FoV variant, the camera FoV is inferred by MoGe-2 and used during optimization.
Lower values are better for Mean, Median, and RMSE, while higher values are better for Acc$_{11.25}$, Acc$_{22.5}$, and Acc$_{30}$.
}
\begin{tabular}{llcccccc}
\toprule
Dataset & Method
& Mean & Median & RMSE
& Acc$_{11.25}$ & Acc$_{22.5}$ & Acc$_{30}$ \\
\midrule

PISR
& MoGe-2
& 12.82 & 9.82 & 17.92 & 0.58 & 0.87 & 0.94 \\

& MoGe-2 + Poppy
& 13.54 & 11.27 & 18.00 & 0.50 & 0.89 & 0.95 \\

& MoGe-2 + Poppy (FoV)
& \textbf{10.87} & \textbf{8.05} & \textbf{16.02}
& \textbf{0.69} & \textbf{0.92} & \textbf{0.96} \\

\bottomrule
\end{tabular}
\label{tab:fov_var}
\end{table}
\noindent In Tab.~\ref{tab:fov_var}, we incorporate the field-of-view (FoV), inferred by MoGe-2, into the Poppy polarization guidance optimization.
Given the predicted FoV, we compute the focal length as $f = \frac{W}{2\tan(\text{FoV}/2)}$, where $W$ is the image width.
Per-pixel viewing directions are then computed by constructing a pinhole camera grid over the image plane.
Specifically, pixel coordinates $(u, v)$ are defined with the origin at the principal point $(c_x, c_y)$, with the $v$-axis pointing upward.
The unprojected ray direction for each pixel is given by
\begin{equation}
    \mathbf{d}(u, v) = \left( \frac{c_x - u}{f_x},\ \frac{c_y - v}{f_y},\ 1 \right),
\end{equation}
where $f_x = f$ and $f_y = f \cdot \frac{H}{W}$ are the horizontal and vertical focal lengths, respectively.
Each direction is then normalized to obtain the unit viewing direction $\hat{\mathbf{d}}(u,v) = \mathbf{d}(u,v) / \|\mathbf{d}(u,v)\|_2$.
Now we compute the elevation angle of a surface with the viewing directions in normals to Stokes conversion, as described in Sec.~\ref{sec:n2s}.

For the PISR dataset, incorporating FoV significantly improves performance, reducing the angular errors and increasing accuracy across all angular thresholds.
This improvement indicates that modeling the camera FoV is important when perspective distortions are present, since the viewing directions used in the polarization model depend on the correct camera geometry.

In scenes with noticeable perspective distortion, assuming an orthographic viewing model can introduce errors in the estimated surface normals.
By explicitly incorporating the FoV, the optimization can better account for the true viewing rays and therefore produce more accurate polarization-consistent normals.
This result suggests that FoV modeling becomes particularly beneficial when objects are close to the camera and perspective effects cannot be neglected.

\section{Effect of varying refractive index}
\begin{table*}[h!]
  \centering
\caption{
\textbf{Comparison of different refractive index $\eta$ settings on the SfPUEL dataset, which contains both dielectric and metallic objects.}
Poppy assumes a refractive index $\eta=1.5$, while the SfPUEL paper uses $\eta=3.2$. 
To analyze sensitivity to the refractive index assumption, we additionally evaluate Poppy with $\eta=3.2$. The best results are highlighted in bold.
}
  \label{tab:eta_var}
  \resizebox{\textwidth}{!}{
    \begin{tabular}{lcccccccccccc}
      \toprule
      & \multicolumn{6}{c}{\textbf{Real}}
      & \multicolumn{6}{c}{\textbf{Synthetic}} \\
      \cmidrule(lr){2-7} \cmidrule(lr){8-13}
      Method
      & Mean & Median & RMSE & Acc11.25 & Acc22.5 & Acc30
      & Mean & Median & RMSE & Acc11.25 & Acc22.5 & Acc30 \\
      \midrule

      MoGe-2
      & 10.84 & 9.87 & 12.85 & 0.60 & 0.94 & 0.98
      & 10.93 & 8.84 & 13.98 & 0.62 & 0.91 & 0.95 \\

      MoGe-2 + Poppy ($\eta=1.5$)
      & \textbf{8.74} & \textbf{7.61} & \textbf{10.91}
      & \textbf{0.75} & \textbf{0.97} & \textbf{0.99}
      & \textbf{9.19} & \textbf{6.84} & \textbf{12.50}
      & \textbf{0.75} & \textbf{0.94} & \textbf{0.96} \\

      MoGe-2 + Poppy ($\eta=3.2$)
      & 8.75 & 7.62 & 10.92
     & \textbf{0.75} & \textbf{0.97} & \textbf{0.99}
      & \textbf{9.19} & \textbf{6.84} & \textbf{12.50}
      & \textbf{0.75} & \textbf{0.94} & \textbf{0.96} \\
      \bottomrule
    \end{tabular}
  }
\end{table*}
\noindent To evaluate the sensitivity of Poppy to refractive index, we run additional experiments using $\eta = 3.2$, which is the value adopted in the original SfPUEL~\cite{sfpuel} work, while our default assumption is $\eta = 1.5$ corresponding to common materials.
As shown in Tab.~\ref{tab:eta_var}, the performance difference between the two settings is negligible across all metrics and scenes, with variations appearing only at the third decimal place in most cases.

Polarization-based normal estimation relates the degree of linear polarization (DoLP) to the elevation angle through Fresnel reflection models as described in Sec.~\ref{sec:n2s}. 
Both diffuse and specular DoLP depend on $\eta$ through rational functions whose variation with respect to $\eta$ is relatively smooth. 
In the range of refractive indices typically encountered in real materials, the change in $\rho_d$ and $\rho_s$ induced by modifying $\eta$ mainly results in small shifts in the DoLP. 

This explains why using $\eta = 1.5$ (as used in Poppy) or $\eta = 3.2$ (as used in the SfPUEL dataset) yields nearly identical optimization behavior and final normal estimates in our experiments.

\section{Comparison of Poppy and direct finetuning}

\noindent Poppy introduces a lightweight test-time guidance approach, which involves just learning per-pixel offsets to the input and output of the backbone network with the backbone weights as frozen. Here, we compare our approach with the alternative of fine-tuning the weights of the backbone estimators, without any learnable offset, using the Stokes loss of the given scene at test-time.

\begin{table}[h!]
\centering
\caption{\textbf{Memory usage comparison between Poppy and direct backbone finetuning.}
Peak memory usage of Poppy and direct finetuning of model weights are reported for each backbone.
Direct finetuning requires gradients and optimizer states for all backbone parameters,
making it infeasible for large models such as Lotus-v2 on a 96\,GB GPU.}
\label{tab:finetune_memory}
\begin{tabular}{lcccc}
\toprule
\textbf{Backbone} & \textbf{Poppy (GB)} & \textbf{Direct Finetuning (GB)} \\
\midrule
Marigold  & 35.12 & 47.9 \\
Lotus-v2  & 66.64 & $>$96 (OOM) \\
MoGe-2    & 15.33 & 21.0 \\
\bottomrule
\end{tabular}
\end{table}

\paragraph{Memory comparison of direct backbone finetuning.}

We compare two settings: our Poppy optimization and 
direct backbone weights finetuning with the same polarization guidance as Poppy in Tab.~\ref{tab:finetune_memory}.
During inference, the backbone performs a forward pass, and only the model weights
and forward activations required for computation must be stored in memory.

Direct finetuning uses more memory demanding because gradients must be
computed for all backbone parameters.
This requires storing intermediate activations for backpropagation and allocating
additional gradient tensors for every model parameter.
As a result, the memory footprint of finetuning is larger than that
of Poppy, especially for large models. In particular, finetuning for Lotus-v2 is not possible because the model could not be trained under the available memory budget (96 GB).

Poppy avoids this issue by keeping the backbone weights frozen and optimizing only
a small set of offset variables (image offsets, normal offsets, and radiance maps).
Since gradients are not computed for the backbone parameters, the backbone can be
executed in inference mode while the optimization updates only the offset tensors.
Consequently, the memory overhead of Poppy is less than direct finetuning.

\begin{table}[h!]
  \centering
  \caption{\textbf{Comparison of Poppy and finetuning.}
  We compare the backbone models (None), direct backbone finetuning with the polarization loss (Finetune), and our polarization-guided optimization (Poppy).
  The best result for each backbone and dataset is highlighted in bold.
  Finetuning results for Lotus-v2 are not reported because the model could not be trained due to GPU memory constraints.}
  \label{tab:finetune_quant}
  \resizebox{0.8\linewidth}{!}{
    \footnotesize
    \setlength{\tabcolsep}{1pt}
    \begin{tabular}{llccccccc}
      \toprule
      \multirow{2}{*}{Backbone} & \multirow{2}{*}{Variant} &
      \multicolumn{4}{c}{Synthetic} &
      \multicolumn{3}{c}{Real} \\
      \cmidrule(lr){3-6}\cmidrule(lr){7-9}
       & & SfPUEL & NeRSP & NeISF & DeepPol
      & SfPUEL & NeRSP & PISR \\
      \midrule

      \multirow{3}{*}{Marigold}
       & None      & 15.19 & 21.47 & 20.03 & 27.29 & 17.45 & 19.03 & 18.06 \\
       & Finetune  & 13.11 & 17.15 & 12.38 & 21.38 & \textbf{11.02} & 17.23 & 17.44 \\
       & Poppy      & \textbf{12.45} & \textbf{16.55} & \textbf{12.33} & \textbf{21.07} & 12.12 & \textbf{17.01} & \textbf{16.71} \\
      \midrule

      \multirow{2}{*}{Lotus-v2}
       & None      & 13.14 & 17.85 & 15.30 & 19.80 & 12.38 & 17.76 & 13.90 \\
       & Poppy      & \textbf{10.22} & \textbf{13.82} & \textbf{9.92} & \textbf{15.08} & \textbf{9.64} & \textbf{15.48} & \textbf{12.82} \\
      \midrule

      \multirow{3}{*}{MoGe-2}
       & None      & 10.93 & 16.08 & 13.55 & 15.95 & 10.84 & 15.63 & \textbf{12.82} \\
       & Finetune  & 9.53 & 12.32 & \textbf{8.64} & \textbf{12.63} & \textbf{8.34} & 15.06 & 15.30 \\
       & Poppy      & \textbf{9.19} & \textbf{11.98} & 9.58 & 12.80 & 8.74 & \textbf{14.51} & 13.54 \\
      \bottomrule
    \end{tabular}
  }
\end{table}

\paragraph{Quantitative comparison with direct finetuning.}

\begin{figure}[h!]
    \centering
    \includegraphics[width=0.5\linewidth]{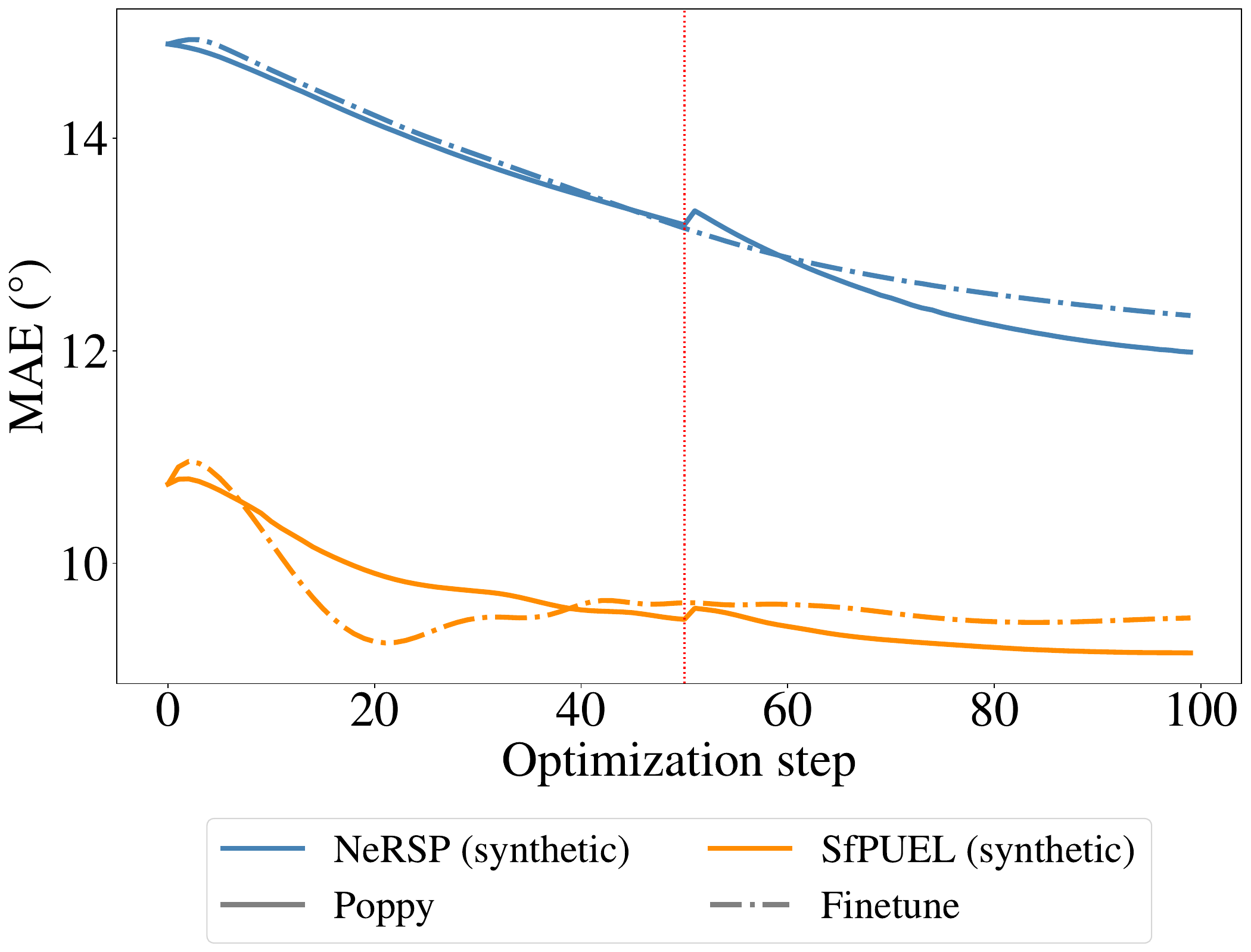}
\caption{
\textbf{MAE convergence curves over optimization steps for Poppy and direct backbone weights finetuning (MoGe-2) across SfPUEL and NeRSP synthetic datasets.}
Poppy converges more smoothly, while finetuning exhibits unstable oscillations despite very small learning rates.
}
    \label{fig:iter_enough}
\end{figure}

Tab.~\ref{tab:finetune_quant} compares the backbone models, direct backbone finetuning with the polarization guidance, and Poppy. The best learning rate is searched and used for each backbone ($10^{-6}$ for Marigold and $10^{-7}$ for MoGe-2) for comparison.
Overall, both finetuning and Poppy achieve similar performance across datasets, with only small differences in MAE. Fig.~\ref{fig:iter_enough} shows that the similar performance is not due to insufficient finetuning steps but rather that both methods converge to comparable solutions.
This indicates that most of the benefits of polarization-guided adaptation can be obtained without updating the backbone weights.

\begin{figure}[h]
    \centering
    \includegraphics[width=0.7\linewidth]{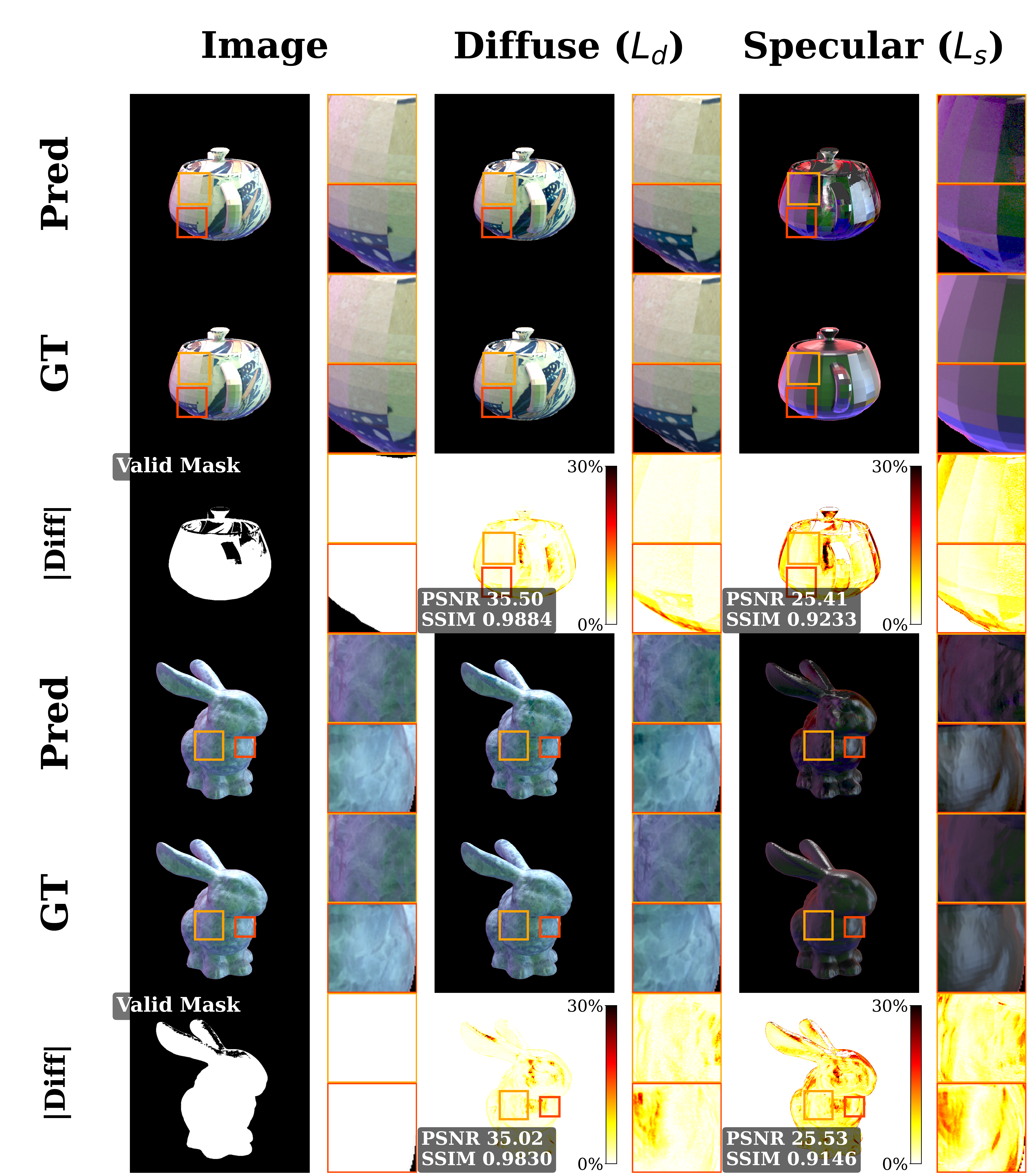}
    \caption{
        \textbf{Qualitative comparison of the predicted diffuse ($L_d$) and specular ($L_s$) radiance components with ground truth~(GT).}
        The absolute error maps (shown as percentages) visualize per-pixel reconstruction error within the valid polarization mask, with overall PSNR and SSIM values reported as labels on each error map.
        The valid mask indicates pixels where polarization-based constraints are reliable and used during optimization.
        The results show that the predicted decomposition closely matches the ground truth across both diffuse and specular regions, with low absolute errors over most valid pixels.
    }
    \label{fig:intrinsic_decomposition}
\end{figure}

\begin{figure}[h]
    \centering
    \includegraphics[width=1.0\linewidth]{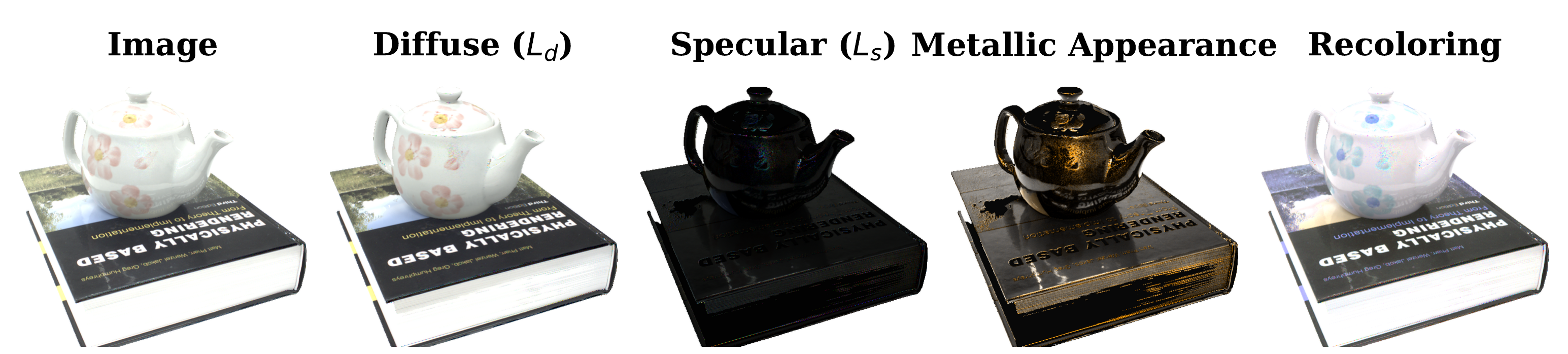}
    \caption{
        \textbf{Applications of radiance decomposition from Poppy for appearance editing.}
        Given the estimated diffuse ($L_d$) and specular ($L_s$) components, different visual effects can be achieved by manipulating each component independently.
        Recoloring modifies the diffuse radiance $L_d$ to change the surface color while preserving specular reflections, while the metallic appearance effect is produced by editing the specular radiance $L_s$.
        These examples demonstrate that the decomposition enables flexible and physically interpretable image editing.
    }
    \label{fig:editing_applications}
\end{figure}

Fig.~\ref{fig:iter_enough} further illustrates the optimization dynamics of both approaches.
While finetuning converges, its loss trajectory is noticeably less stable and oscillates.
In contrast, Poppy converges smoothly, reflecting the well-conditioned nature of optimizing small offset variables against a frozen backbone.

Importantly, Poppy incurs less computational overhead while achieving similar performance to the finetuning approach. As a result, it is more memory efficient and practical for various models. 

\section{Comparison of learned diffuse--specular radiance decomposition}
\noindent Fig.~\ref{fig:intrinsic_decomposition} compares the predicted diffuse and specular radiance components with the corresponding ground-truth decomposition in the NeISF dataset.
The predicted $L_d$ closely reproduces the ground-truth diffuse appearance, preserving the global shading patterns of the objects, while $L_s$ effectively isolates reflective regions, particularly near the object boundaries and corners where strong specular reflections from surrounding surfaces occur.
The absolute error maps confirm low reconstruction error across most valid pixels, and the PSNR and SSIM values further confirm that the decomposition is quantitatively consistent with the ground truth decomposition.

\section{Image editing from our learned decomposition}
\label{sec:image_editing}
\noindent Fig.~\ref{fig:editing_applications} demonstrates appearance editing applications, mentioned in Sec.~\ref{sec:applications}, enabled by the diffuse and specular radiance components learned using Poppy.
Edits applied to the diffuse component $L_d$ modify the surface color appearance while preserving the original specular highlights and reflection patterns, as shown in the recoloring example.
In contrast, edits to the specular component $L_s$ affect the material's reflective properties.
Changing the hue of $L_s$ produces a metallic appearance, simulating materials with strong, colored specular reflections across the entire surface.
By independently controlling $L_d$ and $L_s$, realistic material modifications can be achieved while maintaining consistent lighting and shading.

\end{document}